\documentclass{ecai2014}

\usepackage{rotating}
\usepackage{graphicx}
\usepackage[lined,linesnumbered]{algorithm2e}
\usepackage{amssymb}
\usepackage{amsthm}
\usepackage[table]{xcolor}
\usepackage{hyperref}
\hypersetup{colorlinks,
  citecolor=teal!80!black,
  linkcolor=red!100!black,
  urlcolor=blue}
\usepackage{verbatim}
\usepackage{multirow}
\usepackage{lscape}
\usepackage{fancyhdr}
\usepackage{array}

\newcolumntype{L}[1]{>{\raggedright\let\newline\\\arraybackslash\hspace{0pt}}p{#1}}
\newcolumntype{C}[1]{>{\centering\let\newline\\\arraybackslash\hspace{0pt}}p{#1}}
\newcolumntype{R}[1]{>{\raggedleft\let\newline\\\arraybackslash\hspace{0pt}}p{#1}}

\definecolor{tableHeader}{HTML}{CCCCCC} 
\definecolor{tableShade}{HTML}{CCE3FE}
\theoremstyle{definition} \newtheorem{definition}{Definition}[section]

\newenvironment{theorytable}
{\begin{center}
\begin{tabular}{l C{0.5cm} L{6.5cm} r}}
{\end{tabular}
\end{center}
}

\newenvironment{bridgeRule} 
{
\begin{center}
\begin{tabular}{C{7.5cm} r}}
{\end{tabular}
\end{center}
}

\newenvironment{prooftableonecolumn}
{\begin{center}
\begin{tabular}{L{0.8\columnwidth} r}}
{\end{tabular}
\end{center}
}

 \author{Tomas Trescak \and Carles Sierra \and 
 Simeon Simoff \and \\ Ramon Lopez de Mantaras}
\title{Dispute Resolution Using Argumentation-Based Mediation}

\begin{document}

\maketitle

\begin{abstract} Mediation is a process, in which both parties agree to resolve their dispute by negotiating over alternative solutions presented by a mediator. In order to construct such solutions, mediation brings more information and knowledge, and, if possible, resources to the negotiation table. The contribution of this paper is the automated mediation machinery which does that. It presents an argumentation-based mediation approach that extends the logic-based approach to argumentation-based negotiation involving BDI agents. The paper describes the mediation algorithm. For comparison it illustrates the method with a case study used in an earlier work. It demonstrates how the computational mediator can deal with realistic situations in which the negotiating agents would otherwise fail due to lack of knowledge and/or resources. 
\end{abstract}

\section{Introduction and Motivation}
\label{sec:introduction}

Dispute resolution is a complex process, depending on the will of involved parties to reach consensus, when they are satisfied with the result of negotiation, which allows them to partially or completely fulfil their goals with the available resources. In many cases, such negotiation depends on searching for alternative solutions, which requires an extensive knowledge about the disputed matter for sound argumentation. Such information may not be available to the negotiating parties and negotiation fails. Mediation, a less radical alternative to arbitration, can assist both parties to come to a mutual agreement. This paper presents an argumentation-based mediation system that builds on previous works in the field of argumentation-based negotiation. It is an extension of the work presented in \cite{Parsons1998} and focuses on problems where negotiation stalled and had no solution. In \cite{Parsons1998} agents contain all the knowledge and resources needed to resolve their dispute - a relatively strong assumption in the context of real world negotiations. Agents present arguments, which their opponent can either accept, reject, or they can negotiate on a possible solution. As mentioned earlier, lacking knowledge or resources may lead to an unsuccessful negotiation. In many cases, such knowledge or even alternative resources may be available, but agents are not aware of them.
%; therefore, we propose the role of the mediator, responsible for their discovery and a creation of alternative solutions.

%\usepackage{graphics} is needed for \includegraphics
% \begin{figure}[!h!t]
% \begin{center}
%   \includegraphics[width=0.8\columnwidth]{images/MediationArchitecture}
%   \caption{Architecture of the mediation system}
%   \label{fig:mediation}
% \end{center}
% \end{figure}

% Figure~\ref{fig:mediation} shows our extension of the argumentation-based negotiation system, presented in \cite{Parsons1998}. 

%In this extension, we propose 
Our extension proposes a role of a trust-worthy mediator that possesses extensive knowledge about possible solutions of mediation cases, which it can adapt to the current case. Mediator also has access to various resources that may help to resolve the dispute. Using this knowledge and resources, as well as knowledge and resources obtained from agents, the mediator creates alternative solutions, which become subject to further negotiation.

%In the next section, we summarize previous works in the field of automatic mediation. In Section~\ref{sec:agentArchitecture}
In the next section, we summarise related work in the field of automatic mediation and argumentation-based negotiation. In Section~\ref{sec:agentArchitecture}, we recall the agent architecture proposed by Parsons et al. \cite{Parsons1998} and extend it with the notion of resources for the purposes of the mediation system. Section~\ref{sec:algorithm} presents our mediation algorithm. In Section~\ref{section:example}, we revisit the home improvement agents example from \cite{Parsons1998} and apply our mediation process. Section~\ref{sec:conclusion} concludes this work.

\vspace{-1mm}
\section{Previous Work}
\label{sec:previousWork}
%\vspace{-2mm} 
%!TEX root=document.tex
%\subsection{Computational mediation}
\label{sec:compMed}

\emph{Computational mediation} has recognized the role of the mediator as a problem solver. The {\tt{MEDIATOR}} \cite{Kolodner1989} focused on case-based reasoning as a single-step for finding a solution to a dispute resolution problem. The mediation process was reduced to a one-step case-based inference, aimed at selecting an abstract ``mediation plan''. The work did not consider the value of the actual dialog with the mediated parties. The {\tt{PERSUADER}}  \cite{Sycara1991} deployed mechanisms for problem restructuring that operated over the goals and the relationships between the goals within the game theory paradigm, applied to labor management disputes. To some extent this work is a precursor of another game-theoretic approach to mediation, presented in \cite{Wilkenfeld2004} and the interest-based negotiation approach in \cite{Rahwan2009b}. Notable are recent game-theoretic computational mediators {\tt{AutoMed}} \cite{Chalamish2012} and {\tt{AniMed}} \cite{Lin2011a} for multi-issue bilateral negotiation under time constraints. They operate within known solution space, offering either specific complete solutions ({\tt{AutoMed}}) or incremental partial solutions ({\tt{AniMed}}). Similar to the mediator proposed in the `curious negotiator' \cite{Simoff2002}, both mediators monitor negotiations and intervene when there is a conflict between negotiators. The {\tt{Family\_Winner}} \cite{Bellucci2005} manipulative mediator aimed at modifying the initial preferences of the parties in order to converge to a feasible and mutually acceptable solution. This line of works incorporated ``fairness'' in the mediation strategies \cite{Abrahams2012}.

In real settings information only about negotiation issues is not sufficient to derive the outcome preferences \cite{Visser2011}. An exploratory study \cite{Schei2003} of a multiple (three) issue negotiation setting suggests the need for developing integrative (rather than position-based) negotiation processes which take into account information about the motivational orientation of negotiating parties. Incorporation of information beyond negotiation issues has been the focus of a series of works related to information-based agency \cite{Debenham2004,Debenham2006,Sierra2007}. Value-based argumentation frameworks \cite{Bench-Capon2003}, interest-based negotiation \cite{Rahwan2009b} and interest-based reasoning \cite{Visser2011} considers the treatment of any kind of motivational information that leads to a preference in negotiation and decision making. 

\label{sec:ABN}

In this paper we propose a new mechanism for automatic mediation using \emph{argumentation-based negotiation} (ABN) as a principal framework for mediation. ABN systems evolved from classical argumentation systems, bringing power to agents to resolve potential dispute deadlocks by persuasion of agents in their beliefs and finding common acceptance grounds by negotiation \cite{Sycara1990,Parsons1998,kakas2006,Rahwan2003}. ABN is performed by exchanging arguments, which represent a stance of an agent related to the negotiated subject and constructed from beliefs of the agent. Such a stance can support another argument of the agent, explain why a given offer is rejected, or provide conditions upon which the offer would be accepted. Disputing parties can modify their offer or present a counter-offer, based on the information extracted from the argument. Arguments can be used to attack \cite{Dung1995} other arguments, supporting or justifying the original offer. With certain level of trust between negotiating agents, arguments serve as knowledge exchange carriers \cite{Parsons1998} - here we use such mechanisms to exchange information between negotiating parties and the mediator. The decision of whether to trust the negotiating party or not is a part of the \textit{strategy} of an agent. Different strategies are proposed in \cite{Hadidi2011,Dung2008,Dijkstra2007}. Apart from the strategy, essential are the \textit{reasoning mechanisms} and \textit{negotiation protocols}. Relevant to this work are logic frameworks that use argumentation as the key mechanism for reasoning \cite{Krause1995,Prakken1997,Dung2006,Oren2007}. \textit{Negotiation protocols}, which specify the negotiation procedures include either finite-state machines \cite{Parsons1998}, or functions based on the previously executed actions \cite{Amgoud2007}. The reader is referred to \cite{Rahwan2009} for the recent state-of-the-art in ABN frameworks.

Our ABN framework for mediation allows us to seamlessly design and execute realistic mediation process, which utilises the power of argumentation, using agent logics and a negotiation procedure to search for the common agreement space. We have decided to extend the ABN framework in \cite{Parsons1998}, due to the clarity of its logics. In the next section we recall the necessary aspects of the work in \cite{Parsons1998}. We describe the agent architecture in the ABN systems and define the components that we reuse in our work. Our agents \textit{reason} using argumentation, based on a domain dependent theory specified in a first-order logic. Within the theory, we encode \textit{agent strategies}, by defining their planning steps. Apart from agent theories, \textit{strategy} is defined also in \textit{bridge rules}, explained further in the text. We do not explore a custom protocol, therefore we adopt the one from \cite{Parsons1998}.

% Before getting into the technical aspects, it is worth mentioning that the award of the 2002 Nobel Peace Prize to Jimmy Carter recognizes the role of successful mediation in contemporary world.\footnote{Though Jimmy Carter as a President launched a number of controversial weapons programs, see ``The Nobel Peace Prize 2002 - Presentation Speech.'' [http://www.nobelprize.org/nobel\_prizes/peace/laureates/2002/presentation-speech.html] for the supportive argument about his contribution as a mediator. Whilst Jimmy Carter's mediation between Israel and Egypt (the Camp David Accords) is a well-known classics, less known are his other numerous successes. For example, in 1994  his mediation resulted in a four-month cease-fire agreement in Bosnia at the height of the ethnic violence in the Balkans, and a pledge from all sides to resume peace talks, which eventually led to a peace agreement between Croatia, Bosnia, and Serbia in 1995.  In 2008, his mediation led to the establishment of diplomatic relations between Colombia and Ecuador.} Distinct element of Jimmy Carter's mediation strategies is the intertwining of the settlement-centered strategies, in which the mediator is highly manipulative in order to bring the parties to a resolution, proposed by the mediator, and relationship-centered (transformative) where the mediator assists parties in building mutual trust and understanding, and developing mutually-acceptable solution on their own. Both types of mediation strategies involve elements of argumentation \cite{Raiffa2002}.

\vspace{-2mm}
\section{Agent Architecture}
\label{sec:agentArchitecture}

The ABN system presented in \cite{Parsons1998} is concerned with BDI agents in a multi-context framework, which allows distinct theoretical components to be defined, interrelated and easily transformed to executable components. The authors use different contexts to represent different components of an agent architecture, and specify the interactions between them by means of the bridge rules between contexts. We recall briefly the components of the agent architecture within the ABN system in \cite{Parsons1998} and add a new ``resources'' component for mediation purposes.

\textit{Units} are structural entities representing the main components of the architecture. There are four units within a multi-context BDI agent, namely: the \textbf{C}ommunication unit, and units for each of the \textbf{B}eliefs, \textbf{D}esires and \textbf{I}ntentions. Bridge rules connect units, which specify internal agent architecture by determining their relationship. Three well-established sets of relationships for BDI agents have been identified in \cite{Rao1995}: \textit{strong realism, realism and weak realism}. In this work, we consider strongly realist agents. 
%If these agents do not believe something, they neither desire, nor intend it. 

% Concerning realist agents, if they believe something, they both desire and intend it. Weak realist agents represent a specific case between strong realism and weak realism, explained in detail in \cite{Rao1991}. Figure~\ref{fig:strongRealist} contemplates a strongly realist BDI agent architecture. We provide a detailed listing of bridge rules further in this text.
% 
% %\usepackage{graphics} is needed for \includegraphics
% \begin{figure}[htp]
% \begin{center}
%   \includegraphics[width=0.5\columnwidth]{images/BdiArchitecture}
%   \caption[labelInTOC]{Relations between units corresponding to strong realist BDI agents}
%   \label{fig:strongRealist}
% \end{center}
% \end{figure} 

\textit{Logics} is represented by declarative languages, each with a set of axioms and a number of rules of inference. Each unit has a single logic associated with it. For each of the mentioned B, D, I, C units, we use classical first-order logic, with special predicates B, D and I related to their units. These predicates allow to omit the temporal logic CTL modalities as proposed in \cite{Rao1995}. 
% Parsons et. al added the following axioms to the belief unit to capture such modalities, to satisfy the K, D, 4 and 5 axioms:
% 
% \begin{center}
% \begin{tabular}{p{1cm}l}
% \textbf{K} & $B: B(\varphi \rightarrow \psi) \rightarrow (B(\varphi) \rightarrow B(\psi))$ \\
% \textbf{D} & $B: B(\varphi) \rightarrow \neg B(\neg \varphi)$ \\
% \textbf{4} & $B: B(\varphi) \rightarrow B(B(\varphi))$ \\
% \textbf{5} & $B: \neg B(\varphi) \rightarrow B(\neg B(\varphi))$ \\
% \end{tabular}
% \end{center}
% 
% Desire and intention unit need to satisfy the K, D axioms; thus, the following axioms are added to the desire unit:
% 
% \begin{center}
% \begin{tabular}{p{1cm}l}
% \textbf{K} & $D: D(\varphi \rightarrow \psi) \rightarrow (D(\varphi) \rightarrow D(\psi))$ \\
% \textbf{D} & $D: D(\varphi) \rightarrow \neg D(\neg \varphi)$ \\
% \end{tabular}
% \end{center}
% 
% and to the intention unit:
% 
% \begin{center}
% \begin{tabular}{p{1cm}l}
% \textbf{K} & $I: I(\varphi \rightarrow \psi) \rightarrow (I(\varphi) \rightarrow I(\psi))$ \\
% \textbf{D} & $I: I(\varphi) \rightarrow \neg I(\neg \varphi)$ \\
% \end{tabular}
% \end{center}

\textit{Theories} are sets of formulae written in the logic associated with a unit. For each of the four units, we provide domain dependent information, specified as logical expressions in the language of each unit. 
%In Section~\ref{section:example}, we specify the theory for the home improvement agents and the mediator. 

\textit{Bridge rules} are rules of inference which relate formulae in different units. Following are bridge rules for strongly realist BDI agents:

\vspace{-1mm}
\begin{center}
\begin{tabular}{r c l C{0.1cm} c}
$I: I(\alpha)$ & $\Rightarrow$ & $D:D(\lceil \alpha \rceil)$ & &  $B: \neg B(\alpha)$  $\Rightarrow$  $D: \neg D(\lceil \alpha \rceil)$ \\
$D: \neg D(\alpha)$ & $\Rightarrow$ & $I: \neg I(\lceil \alpha \rceil)$ & & $C: done(e)$  $\Rightarrow$ \ $B: B(\lceil done(e) \rceil)$ \\
$D: D(\alpha)$ & $\Rightarrow$ & $B: B(\lceil \alpha \rceil)$ & & $I: I(\lceil does(e) \rceil)$  $\Rightarrow$  $C:does(e)$ \\
\end{tabular}
\end{center}
\vspace{-1mm}

\textit{Resources} are our extension of the contextual architecture of  strongly realist BDI agents. Each agent can possess a set of resources $R^v$ with a specific importance value for its owner. This value may determine the order in which agents are willing to give up their resources during the mediation process. We define a value function $v: \mathbf{S} \rightarrow \mathbf{R}$, which for each resource $\phi$ specifies a value $\vartheta \in <0,1>$, $v(\phi) = \vartheta$. Set $R^v$ is ordered according to function $v$. 

\textit{Units}, \textit{logics} and \textit{bridge rules} are static components of the mediation system. All participants have to agree on them before the mediation process starts. \textit{Theories} and \textit{resources} are dynamic components, they change during the mediation process depending on the current state of negotiation. 

% Knowing the agent architecture of agents participating in our mediation system, we can now proceed with the explanation of our mediation algorithm. 

\vspace{-1mm}
\section{Mediation Algorithm}
\label{sec:algorithm}

In a mediation process both parties try to resolve their dispute by negotiating over alternative solutions presented by a mediator. Such solutions are constructed, using available knowledge and resources. Agent knowledge is considered private and is not shared with the other negotiating party. Resources to obtain alternative solutions may have a high value for their owners or be entirely missing. Thus, we propose that the role of the mediator is to obtain enough knowledge and resources to be able to construct a new solution. The mediator presents a possible solution to agents (in the form of an \textit{argument}), which they either approve, or reject (\textit{attack}). Parties can negotiate over a possible solution to come to a mutual agreement. Below we formally define the foundations of our algorithm.

% Mediation is an incremental process separated into rounds, during which both parties privately present new knowledge to the mediator. This information contains the description of the problem, point of the view of both agents, their available knowledge and resources to constructs a possible resolution to the current problem. If no such resolution exists, the mediator uses its own resources, knowledge and means of knowledge discovery to search for alternative solution. 

%The mediator presents a possible solution to agents (in the form of an \textit{argument}), which they either approve, or reject (\textit{attack}). Furthermore, parties can negotiate over a possible solution to come to a mutual agreement. Before we proceed with the presentation of our algorithm, we formally define its base part.

\begin{definition}
$\Delta$ is a set of formulae in language $L$. An \textit{argument} is a pair $(\Phi, \omega)$, $\Phi \subseteq \Delta$ and $\omega \in \Delta$ such that: (1) $\Phi \nvdash \perp$; (2) $\Phi \vdash \omega$; and (3) $\Phi$ is a minimal subset of $\Delta$ satisfying 2.
%\begin{enumerate}
%  \item $\Phi \nvdash \perp$
%  \item $\Phi \vdash \omega$
%  \item $\Phi$ is a minimal subset of $\Delta$ satisfying 2.
%\end{enumerate}
\end{definition}

% \begin{definition}
% An argument framework is a pair $(U,R)$ where $U$ is a set of arguments and $R \subseteq U \times U$ is a binary relation over $U$ . Each element of $a \in U$ is called an argument and $aRb$ means that a attacks b. Thus, a is counterargument for b when $aRb$ holds.
% \end{definition}

% Having defined all formal parts of our mediation system, we can now proceed with the presentation of our mediation algorithm.

% , schematically presented in Figure~\ref{fig:algorithm}. 
% 
% %\usepackage{graphics} is needed for \includegraphics
% \begin{figure}[!h!t]
% \begin{center}
%   \includegraphics[width=0.8\columnwidth]{images/MediationAlgorithm}
%   \caption{Architecture of the mediation system}
%   \label{fig:algorithm}
% \end{center}
% \end{figure}

%\subsection{Mediation Game}   
\label{sec:game}

A \emph{mediation game} is executed in one or more rounds, during which both mediator and agents perform various actions in order to resolve the dispute. Algorithm~\ref{algorith:mediation} contemplates our proposal of the mediation game. In the beginning of each round, agents $\alpha$ and $\beta$ have an opportunity to present new knowledge to the mediator $\mu$. This new knowledge is helping their case, or helping to resolve the dispute. Agents can either present knowledge in the form of formulas from their theory or new resources. Resources can be presented in ascending order of importance, one resource in each round or altogether, depending on the strategy of agents. The mediator obtains knowledge by executing function $\Gamma_i^\mu \leftarrow GetKnowledge(i)$, where $i \in \{\alpha, \beta\}$. The mediator incorporates knowledge $\Gamma_i^\mu$ into theory $\Gamma_\mu$, obtaining $\Gamma_\mu^\prime$. Please note, that the belief revision operator $\oplus$ is responsible for eliminating conflicting beliefs from the theory. Using the knowledge in $\Gamma_\mu^\prime$, the mediator tries to construct a new $solution$ by executing the $CreateSolution(\Gamma_\mu^\prime)$ function. If the $solution$ does not exist and agents did not present new knowledge in this round, mediation fails. Therefore, it is of utmost importance that agents try to introduce knowledge in each round. In the next step, the possible outcomes are:

\begin{algorithm}[!h]
\SetKwData{Solution}{solution}
\SetKwData{Result}{nresult}
\SetKwData{Up}{up}
\SetKwFunction{CS}{CreateSolution}
\SetKwFunction{AOK}{Propose}
\SetKwFunction{GK}{GetKnowledge}
\SetKwFunction{NEG}{Negotiate}
\SetKwInOut{Input}{Input}
\SetKwInOut{Output}{Output}
\Input{Agents $\alpha$, $\beta$ and the mediator $\mu$. $\Gamma_\alpha$, $\Gamma_\beta$ and $\Gamma_\mu$ denote the knowledge of $\alpha$, $\beta$ and $\mu$, while $\Gamma_\alpha^\mu$ and $\Gamma_\beta^\mu$ denote the knowledge presented to the mediator $\mu$ respectively  by $\alpha$ and $\beta$. $\oplus$ is a \textit{belief revision operator}}
\Output{Resolution of the dispute, or $\bot$ if solution does not exists.}
\BlankLine
\Repeat{\Solution $\neq \varnothing$}{
        $\Gamma_\alpha^\mu \leftarrow$ \GK($\alpha$)\tcp*[r]{Theory and resources from $\alpha$} 
        $\Gamma_\beta^\mu \leftarrow $ \GK($\beta$)\tcp*[r]{Theory and resources from $\beta$}
        $\Gamma_\mu^\prime \leftarrow \Gamma_\mu \oplus (\Gamma_\alpha^\mu \cup \Gamma_\beta^\mu)$; \\ 
        \Solution $\leftarrow$ \CS($\Gamma_\mu^\prime$); \\
        \If{\Solution = $\bot$ and $\Gamma_\mu = \Gamma_\mu^\prime$} {
                return $\bot$ \tcp*[r]{Missing new knowledge and no solution}}
        \If{\Solution $\neq \bot$} {
                $\langle result_\alpha, \Gamma_\alpha^{\mu^\prime} \rangle \leftarrow \AOK(\mu, \alpha, \Solution)$ \\
                $\langle result_\beta, \Gamma_\beta^{\mu^\prime} \rangle \leftarrow \AOK(\mu, \beta, \Solution)$ \\
                
                $\Gamma_\mu^\prime \leftarrow \Gamma_\mu^{\prime} \oplus (\Gamma_\alpha^{\mu^{\prime}} \cup \Gamma_\beta^{\mu^{\prime}})$ \\             
                \uIf {$\neg result_{\alpha}$ and $\neg result_{\beta}$} 
                {   $\Gamma_\mu^\prime \leftarrow \Gamma_\mu^\prime \oplus \neg \Solution$; \\
                        \Solution $\leftarrow \bot$;
                }
                \ElseIf{$\neg result_{\alpha}$ or $\neg result_{\beta}$}
                {
                        $\langle \Solution^{\prime}$, $\Gamma_\alpha^{\mu^{\prime\prime}}$, $\Gamma_\beta^{\mu^{\prime\prime}} \rangle$ $\leftarrow$ \NEG(\Solution, $\alpha$, $\beta$); \\
                        $\Gamma_\mu^\prime \leftarrow \Gamma_\mu^{\prime} \oplus (\Gamma_\alpha^{\mu^{\prime\prime}} \cup \Gamma_\beta^{\mu^{\prime\prime}})$ \\
                        \eIf{$\neg \Solution^{\prime}$}
                        {
                        $\Gamma_\mu^\prime \leftarrow \Gamma_\mu^{\prime} \oplus (\neg \Solution \cup \neg \Solution^\prime)$ \\
                        \Solution $\leftarrow \bot$; 
                        }
                        {$\Solution \leftarrow \Solution^\prime$}
                }
        }
        $\Gamma_\mu \leftarrow \Gamma_\mu^\prime$;
}
return \Solution
\caption{Mediation algorithm}
\label{algorith:mediation}
\end{algorithm}

\begin{itemize}  
  \item When both agents accept the $solution$, mediation finishes with success.
  \item When both agents reject the $solution$, mediator adds  the incorrect solution $\neg solution$ and the explanation of the rejection $\Gamma_i^{\mu^\prime}$ from both agents to its knowledge $\Gamma_\mu^\prime$ and starts a new mediation round.
  \item When only one agent rejects the solution, a new negotiation process is initiated, where agents try to come to a mutual agreement (e.g. partial division of a specific item) resulting to $solution^\prime$. If this negotiation is successful, the mediator records $solution^\prime$ as a new solution and finishes mediation with a success. If the negotiation fails, the mediator adds the explanation of the failure $\Gamma_i^{\mu^{\prime\prime}}$ and the failed $solution$ to $\Gamma_\mu^\prime$ and starts a new mediation round.
\end{itemize}
%\vspace{-1mm}

The mediation process continues till a resolution is obtained, or fails, when no new solution can be obtained, and no new knowledge can be presented. In the next section, we revisit the example of home improvement agents from \cite{Parsons1998} and apply the mediation algorithm. 
%\vspace{-2mm}
%Below we apply this algorithm to the home improvement agents example in \cite{Parsons1998}.
%In the next section, we evaluate our mediation algorithm, using the home improvement agents problem from \cite{Parsons1998}. 
 
\section{Case Study: Revisiting Home Improvement Agents}
\label{section:example}

%In this section, we revisit the example of home improvement agents from \cite{Parsons1998}, using the mediation algorithm. 
In this example, agent $\alpha$ is trying to hang a picture on the wall. Agent $\alpha$ knows that to hang a picture it needs a nail and a hammer. Agent $\alpha$ only has a screw, and a hammer, but it knows that agent $\beta$ owns a nail. Agent $\beta$ is trying to hang a mirror on the wall. $\beta$ knows that it needs a nail and a hammer to hang the mirror, but $\beta$ currently possesses only a nail, and also knows that $\alpha$ has a hammer. Mediator $\mu$ owns a screwdriver and knows that a mirror can be hung using a screw and a screwdriver. 

The difference with the example in \cite{Parsons1998} is that mediator owns the knowledge and resource needed to resolve the dispute $\mu$ and not the agents. This reflects the reality, when clients seek advice of an expert to resolve their problem. As mentioned in the Section~\ref{sec:agentArchitecture}, agents $\alpha$ and $\beta$ are strongly realist BDI agents using related bridge rules and predicate logic. We now define all the dynamic parts of the mediation system, i.e. domain specific agent theory and bridge rules\footnote{We adopt following notation: A.* is the theory introduced by the agent $\alpha$, B.* is the theory of the agent $\beta$, M.* is the mediator's theory, G.* is the general theory and R.* are bridge rules}.

\subsection{Agent Theories}
\label{sec:caseTheories}

What follows, is the private theory $\Gamma_\alpha$ of the agent $\alpha$, whose intention is to hang a picture:

\begin{theorytable}
I & : & $I_\alpha(Can(\alpha, hang\_picture))$ & (A.1) \\
B & : & $B_\alpha(Have(\alpha, picture))$ & (A.2) \\
B & : & $B_\alpha(Have(\alpha, screw))$ & (A.3) \\
B & : & $B_\alpha(Have(\alpha, hammer))$ & (A.4) \\
B & : & $B_\alpha(Have(\beta, nail))$ & (A.5) \\
B & : & $B_\alpha(Have(X, hammer) \wedge Have(X, nail) \wedge Have(X, picture) \rightarrow Can(X, hangPicture))$ & (A.6)
\end{theorytable}

Please note, that agent $\alpha$, contrarily to the example in \cite{Parsons1998}, no longer knows that a mirror can be hung with a screw and a screwdriver.
%, therefore this belief is not in the theory of agent $\alpha$. 
What follows, is the private theory $\Gamma_\beta$ of agent $\beta$, whose intention is to hang a mirror.

\begin{theorytable}
I & : & $I_\beta(Can(\beta, hangMirror))$ & (B.1) \\
B & : & $B_\beta(Have(\beta, mirror))$ & (B.2) \\
B & : & $B_\beta(Have(\beta, nail))$ & (B.3) \\
B & : & $B_\beta(Have(X, hammer) \wedge Have(X, nail) \wedge Have(X, mirror) \rightarrow Can(X, hangMirror))$ & (B.4) \\
\end{theorytable}

Following is the theory $\Gamma_\mu$ of the mediator $\mu$, related to the home improvement agents case (please note, that mediator's knowledge can consist of many other beliefs, for example learned from other mediation cases):

\begin{theorytable} 
B & : & $B_\mu(Have(\mu, screwdriver))$ & (M.1) \\
B & : & $B_\mu(Have(X, screw)$ $\wedge$ $Have(X, screwdriver)$ $\wedge$ $Have(X, mirror) \rightarrow Can(X,hang\_mirror)).$ & (M.2) \\
B & : & $B_\mu(Have(X, hammer) \wedge Have(X, nail) \wedge Have(X, mirror) \rightarrow Can(X, hangMirror))$ & (M.3) \\
\end{theorytable}

We adopt the following theories from \cite{Parsons1998} with actions that integrate different models reflecting real world processes such as change of ownership, and processes that model decisions and planning of actions. In what follows $i \in \{\alpha, \beta\}$).

\textbf{Ownership.} When an agent (X) gives up artifact (Z) to (Y), (Y) becomes its new owner:

\begin{theorytable} 
B & : & $B_i(Have(X,Z) \wedge Give(X,Y,Z) \rightarrow Have(Y,Z))$ & (G.1) \\
\end{theorytable}

\textbf{Reduction.} If there is a way to achieve an intention, an agent adopts the intention to achieve its preconditions:

\begin{theorytable} 
B & : & $B_i(I_j(Q)) \wedge B_i(P_1 \wedge \dots \wedge P_k \wedge \dots \wedge P_n \rightarrow Q)$ \newline \hspace{3cm} $\wedge \neg B_i(R_1 \wedge \dots \wedge R_m \rightarrow Q) \rightarrow B_i(I_j(P_l))$ & (G.2) \\
\end{theorytable}

\textbf{Generosity} Mediator $\mu$ is willing to give up resource $Q$

\begin{theorytable}
B & : & $B_\mu(Have(\mu, Q)) \rightarrow \neg I_\mu(Have(\mu, Q)).$ & (G.3)
\end{theorytable}

\textbf{Unicity.} When an agent (X) gives an artifact (Z) away, (X) longer owns it:

\begin{theorytable}
B & : & $B_i(Have(X,Z) \wedge Give(X,Y,Z) \rightarrow \neg Have(X,Z))$ & (G.4) \\
\end{theorytable}

\textbf{Benevolence.} When agent i does not need (Z) and is asked for it by X, i will give Z up:

\begin{theorytable}
B & : & $B_i(Have(i,Z) \wedge \neg I_i(Have,i,Z) \wedge Ask(X,i.Give(i,X,Z)) \rightarrow I_i(Give(i,X,Z))) $ & (G.5) \\
\end{theorytable}

\textbf{Parsimony.} If an agent believes that it does not intend to do something, it does not believe that it will intend to achieve the preconditions (i.e. the means) to achieve it:

\begin{theorytable}
B & : & $B_i(\neg I_i(Q)) \wedge B_i(P_1 \wedge \dots \wedge P_j \wedge \dots \wedge P_n \rightarrow Q) \rightarrow \neg B_i(I_i(P_j))$ & (G.6) \\
\end{theorytable}

\textbf{Unique choice.} If there are two ways of achieving an intention, only one is intended. Note that we use $\triangledown$ to denote exclusive or.

\begin{theorytable}
B & : & $B_i(I_i(Q)) \wedge B_i(P_1 \wedge \dots \wedge P_j \wedge \dots \wedge P_n \rightarrow Q)$ \newline \hspace{3cm} $\wedge B_i(R_1 \wedge \dots \wedge R_n \rightarrow Q) \rightarrow $ \newline \hspace{1cm} $ B_i(I_i(P_1 \wedge \dots \wedge P_n)) \triangledown B_i(I_i(R_1 \wedge \dots \wedge R_n))$ & (G.7) \\
\end{theorytable}

% \paragraph{Trust in mediator} Whatever mediator $\mu$ believes, any other agent $X$ believes as well.
% \begin{theorytable} 
% B & : & $B_\mu(\varphi) \rightarrow B_X(\varphi).$ & (G.7)
% \end{theorytable}

A theory that contains free variables (e.g. X) is considered the \textit{general theory}, while a theory with bound variables (e.g. $\alpha$ or $\beta$) is considered the \textit{case theory}. The mediator stores only the \textit{general theory} for its reuse with future cases. In addition, an agent's theory contains rules of inference, such as modus ponens, modus tollens and particularization. 

%Apart from the above mentioned, an agent's theory also contains rules of inference, such as modus ponens, modus tollens and particularization. 

\subsection{Bridge Rules}

What follows, is a set of domain dependent bridge rules that link inter-agent communication and the agent's internal states.

\textbf{Advice.} When the mediator $\mu$ believes that it knows about possible intention $I_X$ of $X$ it $tells$ it to $X$. Also, when mediator $\mu$ knows something ($\phi$) that can help to achieve intention $\varphi$ of agent $X$, mediator $tells$ it to $X$.

\begin{bridgeRule}
$B_\mu(I_X(\varphi)) \Rightarrow Tell(\mu,X, B_\mu(I_X(\varphi)))$ & (R.1) \\
$B_\mu(I_X(\varphi)) \wedge B_\mu(\phi \to I_X(\varphi)) \Rightarrow Tell(\mu,X, B_\mu(\phi \to I_X(\varphi)))$ & (R.2)
\end{bridgeRule}

\textbf{Trust in mediator} When an agent (i) is told of a belief of mediator ($\mu$), it accepts that belief:
\begin{bridgeRule}
% $B: B_\mu(I_X(\varphi)) \Rightarrow C: Tell(\mu,X,B_\mu(I_X(\varphi)))$. & (R.3.1) \\
%$B: B_\mu(\varphi) \Rightarrow C: Tell(\mu,X,B_\mu(\varphi))$. & (R.3.1) \\
% $B: \neg B_\mu(I_X(\varphi)) \Rightarrow C: Tell(\mu,X,\neg B_\mu(I_X(\varphi)))$. & (R.3.3) \\
%$B: \neg B_\mu(\varphi) \Rightarrow C: Tell(\mu,X,\neg B_\mu(\varphi))$. & (R.3.2) \\
$C: Tell(\mu, i, B_\mu(\varphi)) \Rightarrow B : B_i(\varphi).$ & (R.3) \\
%$C: Tell(\mu, i, \neg B_\mu(\varphi)) \Rightarrow B : \neg B_i(\varphi).$ & (R.3.4) 
\end{bridgeRule}

\textbf{Request.} When agent (i) needs (Z) from agent (X), it asks for it:
\begin{bridgeRule}
$I:I_i(Give(X,i,Z)) \Rightarrow  C: Ask(i,X,Give(X,i,Z)).$ & (R.4)
\end{bridgeRule}

\textbf{Accept Request.} When agent (i) asks something (Z) from agent (X), and it is not in intention of (X) to have (Z), it is given to i:
\begin{bridgeRule}
$C: Ask(i,X,Give(X,i,Z)) \wedge \neg I_X(Have(X,Z)) \Rightarrow I_i(Give(X,i,Z)).$ & (R.5)
\end{bridgeRule}

% \paragraph{Awareness of intentions.} Agents are aware of their intentions.
% \begin{bridgeRule}
% $I : I_i(\alpha) \Rightarrow B : B_i(I_i(\alpha)).$ & (R.6) \\
% $I : \neg I_i(\alpha) \Rightarrow B : B_i(\neg I_i(\alpha)). $ & (R.7)
% \end{bridgeRule}
 
% \paragraph{Offer.} When an agent (i) has the intention of offering something (Z) to another agent
% (X), it informs the recipient of this fact:
% \begin{bridgeRule}
% $I : I_i(Give(i, X, Z)) \Rightarrow  C : Tell(i, X, Give(i,X,Z)).$ & (R.8)
% \end{bridgeRule}
% 
% \paragraph{Impulsiveness.} When an agent believes it has an intention, it adopts that intention.
% \begin{bridgeRule}
% $B : B_i(I_i(\alpha)) \Rightarrow I : I_i(\alpha).$ & (R.6)
% \end{bridgeRule}

\subsection{Resources}

% Agents' theories from Section~\ref{sec:caseTheories} specify that agent $\alpha$ owns a picture, a screw and a hammer. 
In Section~\ref{sec:agentArchitecture}, we have introduced the notion of importance of resources, which defines the order in which agents are giving up their resources during the mediation process. The picture and the hammer depend on the successful accomplishment of agent's $\alpha$ goal and have an importance value of 1.  Agent $\beta$ owns a mirror and a nail, both with importance 1. All other resources have importance 0.

\subsection{Argumentation System}

Our automatic mediation system uses the ABN system, proposed in \cite{Parsons1998}, which is based on the one proposed in \cite{Krause1995}. The system constructs a series of logical steps (arguments) for and against propositions of interest and as such may be seen as an extension of classical logic. In classical logic, an argument is a sequence of inferences leading to a true conclusion. It is summarized by the schema $\Gamma \vdash (\varphi, G)$,
where $\Gamma$ is the set of formulae available for building arguments, $\vdash$ is a suitable consequence relation, $\varphi$ is the proposition for which the argument is made, and $G$ indicates the set of formulae used to infer $\varphi$, with $G \subseteq \Gamma$.

\subsection{Mediation}

In this section, we follow Algorithm~\ref{algorith:mediation} and explain how we can resolve the home improvement agent dispute using automatic mediation. In comparison to Parsons et al. \cite{Parsons1998}, our agents do not possess all the knowledge and resources to resolve their dispute; thus the classical argumentation fails. The mediation algorithm runs in rounds and finishes with:

\begin{enumerate}
  \item Success, when both agents accept the solution proposed by the mediator.
  \item Failure, when the mediator can not create a new solution and no new knowledge or resources are presented in two consecutive rounds.
\end{enumerate}

The algorithm starts with the mediator gathering information about the dispute from both agents (function \textit{GetKnowledge}). In the first round, agents $\alpha$ and $\beta$ state their goals, which become part of the mediator's  beliefs $B_\mu$:

\begin{theorytable}
B & : & $B_\mu(I_\alpha(Can(\alpha, hang\_picture)))$ & (M.4) \\
B & : & $B_\mu(I_\beta(Can(\beta, hangMirror)))$ & (M.5) \\
\end{theorytable}

With this new theory, the mediator tries to construct a new solution, and it fails. Therefore, in the next round, agents have to present more knowledge or resources. Failing to do so would lead to failure of the mediation process. To speed things up, we assume that agents presented all the necessary knowledge and resources in this single step, although this process can last several rounds depending on the strategy of an agent. For example, if a ``cautious'' agent owns more than one resource, it chooses to give up the resource with the lowest importance. 

\begin{center}

\begin{theorytable}
B & : & $B_\mu(Have(\alpha, picture))$ & (M.6) \\ 
B & : & $B_\mu(Have(\alpha, screw))$ & (M.7) \\
B & : & $B_\mu(Have(\alpha, hammer))$ & (M.8)  \\
B & : & $B_\mu(Have(\beta, nail))$ & (M.9) \\
B & : & $B_\mu(Have(\beta, mirror))$ & (M.10) 
\end{theorytable}

\end{center}

With this new information, the mediator is finally able to construct the {\tt{solution}} to the dispute consisting of three different arguments. With the following two arguments, mediator proposes agent $\beta$ to hang the mirror using the screw and the screwdriver (M.2), and screw can be obtained from the agent $\alpha$ and the screwdriver obtained from the mediator itself (Please note, that this knowledge is part of the support for the presented arguments). The first argument is: $(I_\beta(Give(\alpha,\beta,screw)), P_\beta^\prime)$, where $P_\beta^\prime$ is\footnote{$mp$ stands for \textit{modus ponens} and $pt$ stands for \textit{particularization}}:

\begin{prooftableonecolumn}
\{(M.2),(M.5),(G.2)\}  $\vdash_{pt,mp}$ \newline \hspace{0.5cm} $B_\mu(I_\beta(Have(\beta,screw)))$ & (M.11) \\
\{(M.7),(G.1)\}  $\vdash_{mp}$ \newline \hspace{0.5cm}  $B_\mu(Give(\alpha, Y, screw) \rightarrow Have(Y, screw))$ & (M.12) \\
\{(M.11),(M.12),(G.2)\}  $\vdash_{pt,mp}$ \newline \hspace{0.5cm} $B_\mu(I_\beta(Give(\alpha,\beta,screw)))$ & (M.13) \\
\{(M.13)\}  $\vdash_{R.1}$  $Tell(\mu,\beta, I_\beta(Give(\alpha,\beta,screw)))$ & (M.14) \\
\{(M.14)\}  $\vdash_{R.3}$ $I_\beta(Give(\alpha,\beta,screw))$ & (M.15) \\
\end{prooftableonecolumn} 

The second argument is: $(I_\beta(Give(\mu,\beta,screwdriver)),P_\beta^{\prime\prime})$, where $P_\beta^{\prime\prime}$ is

\begin{prooftableonecolumn}
\{(M.2),(M.5),(G.2)\} $\vdash_{pt,mp}$ \newline \hspace{0.5cm} $B_\mu(I_\beta(Have(\beta,screwdriver)))$ & (M.16) \\
\{(M.1),(G.1)\} $\vdash_{mp}$ \newline \hspace{0.3cm} $B_\mu(Give(\mu, Y, screwdriver) \rightarrow Have(Y, screwdriver))$ & (M.17) \\
\{(M.16),(M.17),(G.2)\} $\vdash_{pt,mp}$ \newline \hspace{0.5cm} $B_\mu(I_\beta(Give(\mu,\beta,screwdriver)))$ & (M.18) \\
\{(M.18)\} $\vdash_{R.1}$ \newline \hspace{0.5cm} $Tell(\mu,\beta, I_\beta(Give(\mu,\beta,screwdriver)))$ & (M.19) \\
\{(M.19)\} $\vdash_{R.3}$ \newline \hspace{0.5cm} $I_\beta(Give(\mu,\beta,screwdriver))$ & (M.20) \\
\end{prooftableonecolumn}

These two arguments represent advices to $\beta$ on how it can achieve its goal (B.1) that was communicated to mediator $\mu$ as (M.5). Using bridge rule (R.4) $\beta$ converts this into the following actions:

\begin{center}
\{M.15\} $\vdash_{R.4} Ask(\beta,\alpha, Give(\alpha,\beta, screw)).$ \\
\{M.20\} $\vdash_{R.4} Ask(\beta,\mu, Give(\mu,\beta, screwdriver)).$
\end{center}

When both $\alpha$ and $\mu$ receive this request, they convert this into \textit{accept request} action using bridge rule (R.5). Mediator accepts this request due to the \textit{generosity} theory (G.3), which defines that it is not an intention of mediator to own anything. Agent $\beta$ cannot find a counter-argument that would reject this request (it does not need the nail) and accepts it. With the screw, the screwdriver, the mirror and knowledge on how to hang the mirror using these tools, $\beta$ can fulfil its goal, and it no longer needs the nail. Therefore, the following argument that solves the goal of $\alpha$ is also accepted: $(I_\alpha(Give(\beta,\alpha,nail)),P_\alpha)$, where $P_\alpha$ is:

% \begin{center}
% $(I_\alpha(Give(\beta,\alpha,nail)),P_\alpha)$
% \end{center}
% where $P_\alpha$ is\footnote{$mp$ stands for \textit{modus ponens}}:
% 
\begin{prooftableonecolumn}
\{(M.3),(M.4),(G.2)\} $\vdash_{mp}$ $B_\mu(I_\alpha(Have(\alpha,nail)))$ & (M.21) \\
\{(M.9),(G.1)\} $\vdash_{mp}$  \newline \hspace{0.5cm} $B_\mu(Give(\beta, Y, nail) \rightarrow Have(Y, nail))$ & (M.22) \\
\{(M.21),(M.22),(G.2)\}  $\vdash_{pt,mp}$  \newline \hspace{0.5cm} $B_\mu(I_\alpha(Give(\beta,\alpha,nail)))$ & (M.23) \\
\{(M.23)\} $\vdash_{R.1}$  \newline \hspace{0.5cm} $Tell(\mu,\alpha, B_\mu(I_\alpha(Give(\beta,\alpha,nail))))$ & (M.18) \\
\{(M.18)\} $\vdash_{R.3}$ $I_\alpha(Give(\beta,\alpha,nail))$ & (M.19) \\
\end{prooftableonecolumn}

we convert this into action, using the bridge rule $R.1$ into:

\begin{center}
\{M.19\} $\vdash_{R.1} Ask(\alpha,\beta, Give(\beta,\alpha, nail)).$
\end{center}

When agent $\beta$ receives this request, $\beta$ can accept it by the bridge rule (R.5). This is only possible because of the previous two arguments, when an alternative plan to hang the mirror was presented to $\beta$, otherwise $\beta$ would not be willing to give up the nail needed for his plan. Agent $\beta$ can now decide between two plans using (G.7); therefore it decides to \textit{give} $\alpha$ the nail and both agents were able to fulfil their goals (we assume that $\beta$ does not want to sabotage the mediation).

\section{Conclusion}
\label{sec:conclusion}
%!TEX root=document.tex

% In comparison with previous work, we allow existence of thinner agents and a thick mediation entity, that can use its extensive knowledge to create alternative solution to disputes. Moreover, we have introduced the concept of resources, which allows two level inference on possible solutions concerning resource disputes. Figure \ref{fig:mediation} contemplates the architecture of our mediation system, portrayed as an extension of the system presented in \cite{Parsons1998}. 

%There are many reasons why negotiation may end up in a need for mediation. Frequently, in real settings, the information related to negotiated issues available to the negotiators may not be complete and sufficient for the decision-making. 

Mediation brings more information and knowledge to the negotiation table, hence, an automated mediator would need the machinery that could do that. Addressing this issue in an automated setting, we have presented an ABN approach that extends the logic-based approach to ABN involving BDI agents, presented in \cite{Parsons1998}. We have introduced a mediator in the multiagent architecture, which has extensive knowledge concerning mediation cases and access to resources. Using both, knowledge and resources, the mediator proposes solutions that become the subject of further negotiation when the agents in conflict cannot solve the dispute by themselves. We have described our mediation algorithm and illustrated it with the same case study introduced in \cite{Parsons1998}. The presence of a mediator in ABN allows to deal with realistic situations when negotiation is stalled. In this work we assumed that the agents and the mediator operate within the same ontology, describing the negotiation domain. In real settings, the negotiators may interpret the same term differently. In order to avoid this, mediation will require the initial alignment of the ontologies with which all parties operate.  

%We have situated our approach within the contexts of argumentation-based negotiation and of computational mediation. 
%We advocate that this is more realistic than letting the agents try to solve disputes alone.
%We believe that the presence of a mediator is of interest in ABN because it allows to deal with realistic situations in which the negotiating agents would otherwise fail due to lack of knowledge and/or resources. 
%We plan to develop an elegant alignment process for respective ontologies based on argumentation between the agents and the mediator within the same ABN framework.  

%\vspace{-1mm}
\bibliographystyle{ecai2014} 
\bibliography{bibliography}

\end{document}